\def\year{2018}\relax
\documentclass[letterpaper]{article} 
\usepackage{aaai18}  
\usepackage{times}  
\usepackage{helvet}  
\usepackage{courier}  
\usepackage{url}  
\usepackage{graphicx}  
\usepackage{multirow}  
\begin{document}

\title{Generating Music Medleys via Playing Music Puzzle Games}
\author{Yu-Siang Huang$^\dagger$, Szu-Yu Chou$^{\dagger\ddagger}$, Yi-Hsuan Yang$^\dagger$\\
$^\dagger$Research Center for IT innovation, Academia Sinica, Taiwan\\
$^\ddagger$Graduate Institute of Networking and Multimedia, National Taiwan University, Taiwan\\
}

\maketitle
\begin{abstract}
Generating music medleys is about finding an optimal permutation of a given set of music clips.
Toward this goal, we propose a self-supervised learning task, called the music puzzle game, to train neural network models to learn the sequential patterns in music. 
In essence, 
such a game requires machines to correctly sort a few multisecond music fragments.
In the training stage, we learn the model by sampling multiple non-overlapping fragment pairs from the same songs and seeking to predict whether a given pair is consecutive and is in the correct chronological order.
For testing, we design a number of puzzle games with different difficulty levels, the most difficult one being music medley, which requiring sorting fragments from different songs.
On the basis of state-of-the-art Siamese convolutional network, we propose an improved architecture that learns to embed frame-level similarity scores computed from the input fragment pairs to a common space, where fragment pairs in the correct order can be more easily identified. Our result shows that the resulting model, dubbed as the similarity embedding network (SEN), performs better than competing models across different games, including music jigsaw puzzle, music sequencing, and music medley.
Example results can be found at our project website, \url{https://remyhuang.github.io/DJnet}.
\end{abstract}

\section{Introduction}

Recent years have witnessed a growing interest in unsupervised methods for sequential pattern learning, notably in the computer vision domain. This can be approached by the so-called \emph{self-supervised learning}, which exploits the inherent property of data for setting the learning target. For example, 
\cite{misra2016shuffle}, \cite{fernando2016self} and
\cite{lee2017unsupervised} leveraged the temporal coherence of video frames as a supervisory signal and formulated representation learning as either an 
order verification or a sequence sorting task. \cite{lotter17iclr}, on the other hand, explored prediction of future frames in a video sequence as the supervisory signal for learning the structure of the visual world. These prior arts demonstrate that learning discriminative visual features from massive unlabeled videos is possible.

\begin{figure}[tb]
\includegraphics[width=.32\columnwidth]{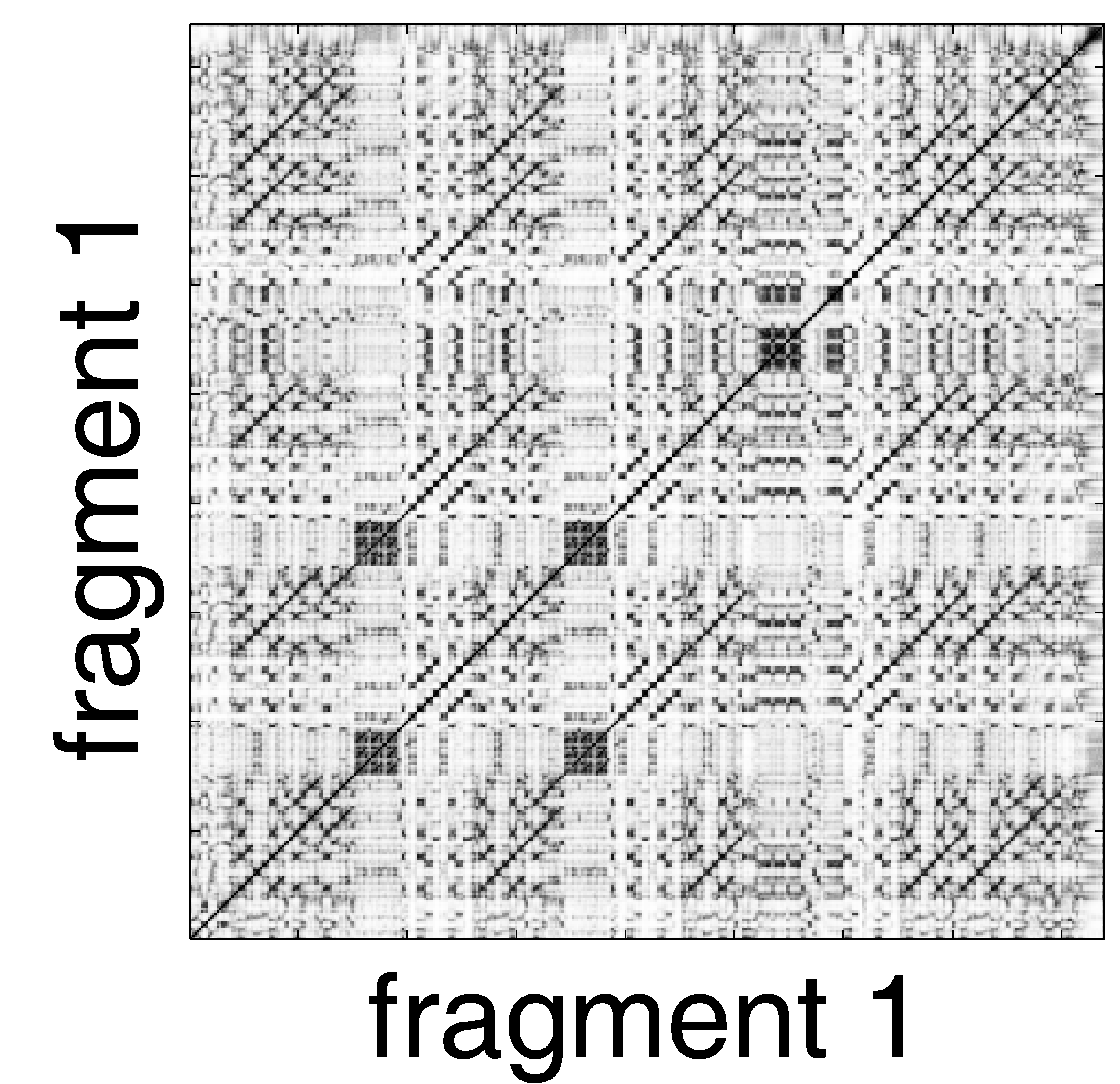}
\includegraphics[width=.32\columnwidth]{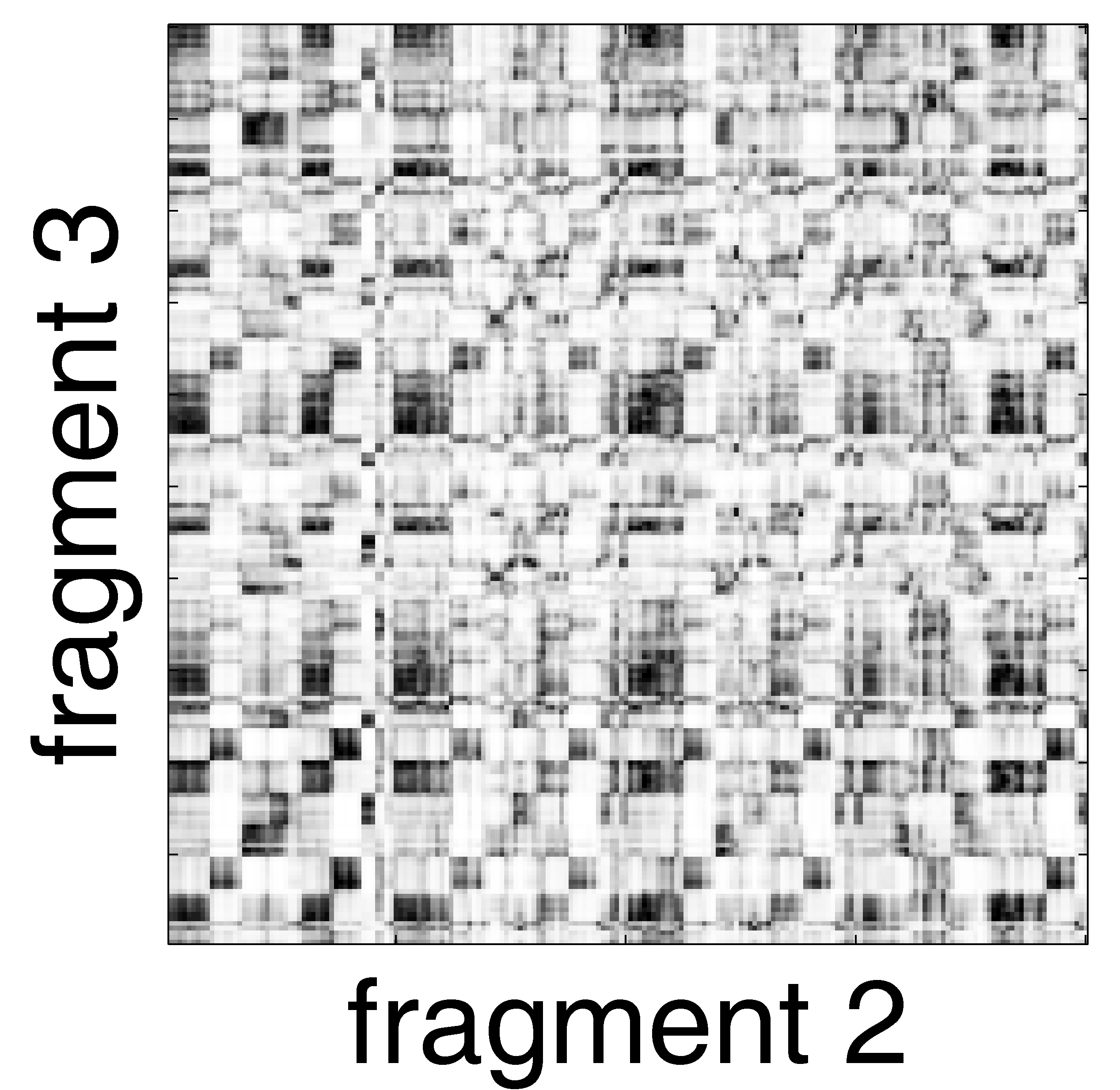}
\includegraphics[width=.32\columnwidth]{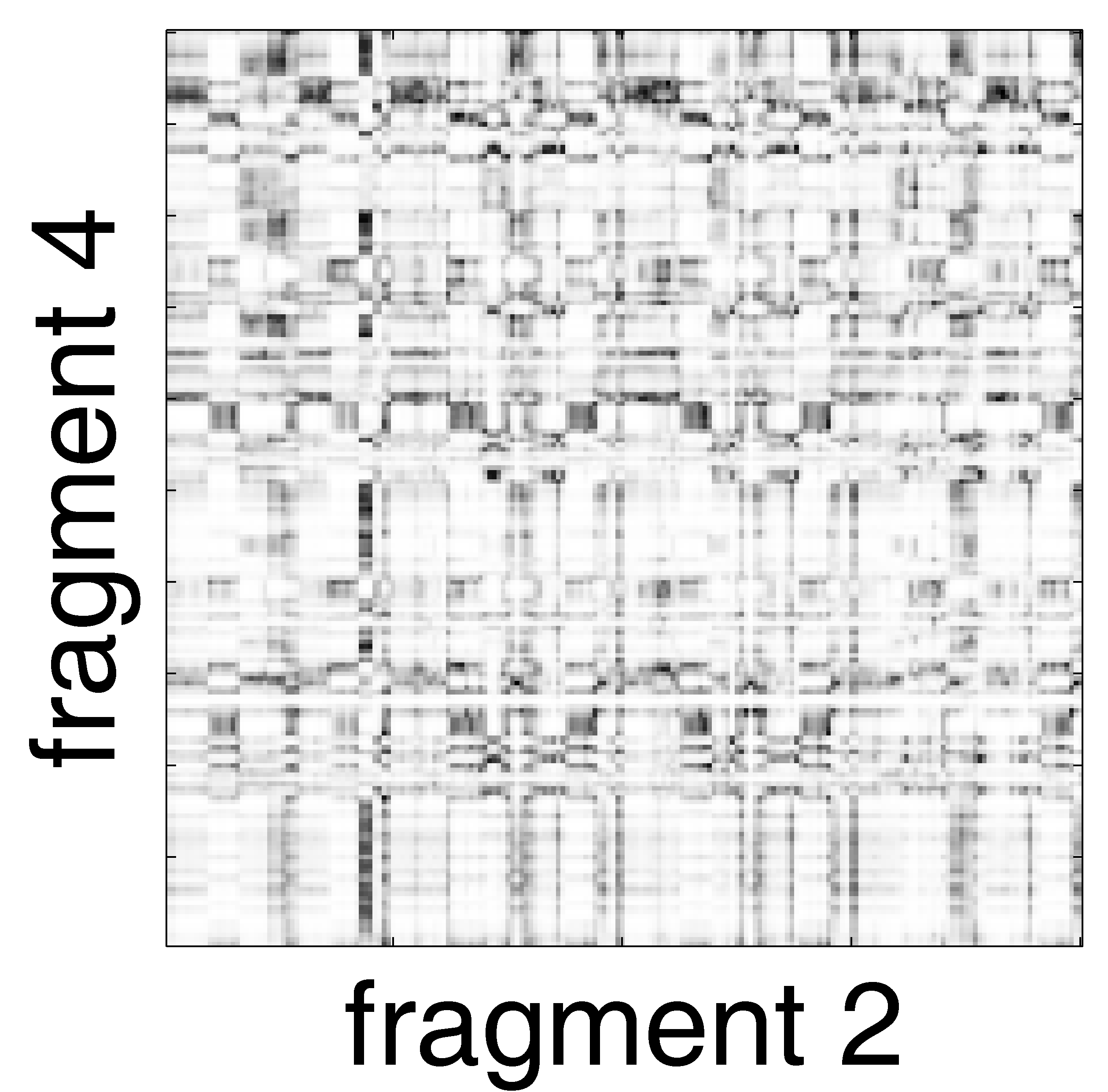}\\
(a)~~~~~~~~~~~~~~~~~~~~~~~~~~(b)~~~~~~~~~~~~~~~~~~~~~~~~~~~~~(c)
\centering
\caption{Similarity matrix between: (a) two identical music fragments, (b) fragments from a song and its cover version (i.e. the same song but different singer), (c) fragments from two different songs of the same singer. The goal of the proposed network is to learn patterns from such matrices.}
\label{fig:sim}
\end{figure}

From a technical standpoint, this paper studies how such a self-supervised learning methodology can be extended to audio, which has been less attempted. In particular, we focus on learning from music sequences. Music is known for its multilevel, hierarchical organization, with higher-level building blocks made up of smaller recurrent patterns \cite{widmer16tist,hudson2011musical}.
While listening to music, human beings can discern those patterns, make predictions on what will come next, and hope to meet their expectations. Asking machines to do the same is interesting on its own, and it poses interesting challenges that do not present, or have not been considered, in the visual domain. 

First, the input instances to existing models are usually video frames (which are images) sampled from each video sequence. Each frame can be viewed as a snapshot of a temporal moment, and the task is to correctly order the frames per video. In contrast, meaningful basic unit to be ordered in music has to be an audio sequence itself. Therefore, the input instances in our case are multisecond, non-overlapping music fragments,
which have a temporal dimension.

Second, existing works in the visual domain considered at most four frames per video \cite{lee2017unsupervised}, mainly due to the concern that the possible permutations increase exponentially along with the number of sampled frames.  However, as a song is typically a few minutes long, we consider up to ten (multisecond) music fragments per song.



Lastly, while it makes less sense to order video frames sampled from different video sequences, for music it is interesting and practically useful if we can find an ordering of a bag of music fragments sampled from different songs. Indeed, music fragments of different songs, when properly ordered, can be listened to consecutively with pleasure \cite{lin15tomm}, given that every pair of consecutive fragments share or follow some harmonic or melodic patterns. For example, Disc Jockeys (DJs) are professional audio engineers who can nicely perform such a \emph{music medley generation} task. Therefore, we also include this task to evaluate the performance of our model. In doing so, from an application standpoint, this paper also contributes to the realization of automatic music medley generation, an important step toward making an AI DJ \cite{djnet}.
The hope is that one day AI would possess certain level of music connoisseurship and can serve as a DJ to create music remixes or mashups professionally.

\begin{figure*}[tb]
\centering
 \includegraphics[width=\textwidth]{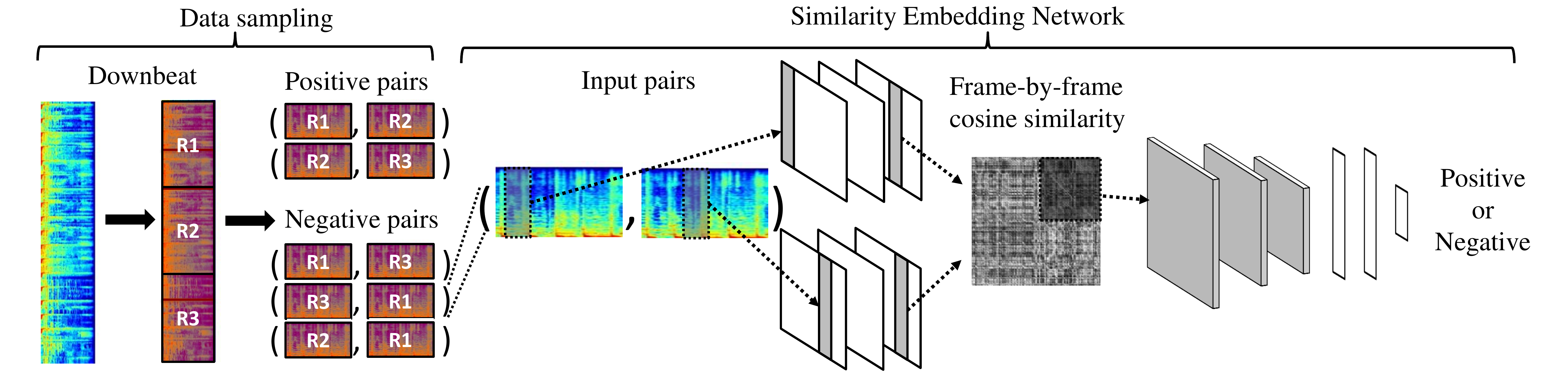}
\caption{Illustration of the proposed similarity embedding network and its application to solving music jigsaw puzzle.}
\label{fig:flow}
\end{figure*}

Drawing the analogy that a fragment is like a puzzle piece, we propose to refer to the task of assembling multiple music fragments in proper order as the \emph{music puzzle games}. 
Similar to previous work in the visual domain, we exploit the temporal coherence of music fragments as the supervisory signal to train our neural networks via a music puzzle game. What's different is that we differentiate four aspects of a music puzzle game and investigate the performance of our models with different types of games. The four aspects are: 1) number of fragments to be ordered, 2) temporal length of the fragments (whether the length is fixed or arbitrary), 3) whether there is a clear cut at the boundary of fragment pairs, and 4) whether the fragments are from the same song or not. For example, other than uniformly sample a song for fragments, we also employ downbeat tracking \cite{bock2016madmom} to create musically meaningful fragments \cite{upham2013coordination}. 

In view of the second challenge mentioned above, we propose to take fragment pairs as input to our neural network models and lastly use a simple heuristic to decide the final ordering. For a music puzzle game with $n$ fragments, this pair-wise approach requires our models to evaluate in total $2\cdot{{n}\choose{2}}=n(n-1)$ pairs, which is much fewer than the $n!$ number of possible permutations and accordingly opens up the possibility to consider $n>4$ fragments.


Moreover, in view of the first challenge mentioned above, we propose an novel model, called the similarity embedding network (SEN), to solve the music puzzle games. The main idea is to compute frame-level similarity scores between each pair of short-time frames from the two input fragments, and then learn to embed the resulting similarity matrix \cite{serra12aaai} into a common space, where coherent and incoherent fragment pairs can be more easily distinguished.

The idea of learning from the similarity matrices has roots in the so-called recurrence plot \cite{marwan2007recurrence}, which provides a way to visualize the periodic nature of a trajectory (i.e. a time-series data) through a phase space. 
Given two trajectories, we can similarly compute their point-by-point (or frame-by-frame) similarity to mine patterns from the resulting matrix.
For our tasks, learning from the similarity matrices is promising, for we can examine temporal correspondence between fragments in more details, as suggested by the example similarity matrices shown in Figure \ref{fig:sim}. Our experiments show that SEN performs consistently better than competing models across different music puzzle games.

\section{Music Puzzle Games}

\subsection{Background and Significance}

In academia, some researchers have investigated the design of music-based puzzle games, mostly for education purposes. 
A notable example is the work presented by \cite{hansen2013music}, which experimented with a number of designs of sound-based puzzle game to train the listening abilities of visually impair people. A 
music clip was divided into several fragments and a player had to rearrange them in order to reconstruct the original song. For advanced players, they further applied pitch and equalization shifts randomly on fragments, requiring the players to detect those transpositions to complete the puzzle. However, in this work a music clip was divided into pieces at arbitrary timepoints. This way, there may be no clear cut at the boundary of the fragments, providing strong temporal cues that make the game easier: when the fragments are in incorrect order, the result will not only sound unmusical but also unnatural.

More recently, \cite{smith2017crosssong} improved upon this wok by dividing songs at downbeat positions, which often coincide with chord changes \cite{bock2016joint} and provides clearer cut among the fragments. Moreover, based on a ``mashability'' measure \cite{davies2014automashupper}, they proposed an algorithm to create cross-song puzzles for more difficult games. They claimed that the game can train the musical and logical reasoning of ordinary people.

Although these previous works are interesting, their focus is on the design of the puzzle games for human beings, rather than on training machines to solve such games. In contrast, we let machine learn sequential patterns (and logic) in the musical world in a self-supervised learning manner by playing and solving such games. 

Another benefit of experimenting with the music puzzle games is that the input to such games are sequences (not images). Therefore, similar network architecture may be applied to time series data in other domains as well.



In what follows, we firstly discuss the design of music puzzle games for machines, and then present a mathematical formulation of the learning problem.

\begin{table}[tb]
\centering
\caption{Characteristics of the music fragments that are to be ordered by our model in various music puzzle games}
\label{table:aspect}
\begin{tabular}{|l|lll|}
\hline
 &Jigsaw Puzzle &Sequencing &Medley \\
\hline
\hline
number &3, 4, 6, 8 &10 &7--11\\
length &fixed~/~arbitrary &arbitrary &arbitrary \\
boundary & unclear~/~clear &clear &clear \\
from &same song &same song &cross song \\
\hline
\end{tabular}
\end{table}

\subsection{Game Design}

As shown in Table \ref{table:aspect}, we consider four aspects in designing the music puzzle games. First, the number of fragments $n$ to be ordered; larger $n$ implies more computational cost and a more difficult game, as potentially more orderings of the fragments would look plausible.
Second, whether the length of the fragments in a game is the same. Our machine model has to deal with input sequences of arbitrary length, if the length of the fragments is not fixed. 
Third, whether there is a clear cut at the boundary of fragment pairs. Arbitrary segmentation of a song may lead to obvious continuity at the boundary of two fragments and make the game too easy. It is conceivable that when the boundaries are clearer, the puzzle game will be more difficult. 
Fourth, whether the fragments are from the same song or not. 

As also shown in Table \ref{table:aspect}, we consider three different music puzzle games in this paper, with progressively increasing level of difficulty.
For \textbf{music jigsaw puzzle}, we create fragments by dividing a 24-second music clip at either equally-spaced or at downbeat-informed timepoints. Because we are interested in comparing the performance of the proposed SEN model against those proposed to solve video puzzles, we vary the value of $n$ from 3 to 8 in this game.

The second game is more difficult in that the fragments are taken from a \emph{whole song}. Moreover, each fragment represents a \emph{section} of the song, such as the intro, verse, chorus and bridge \cite{paulus10ismir,MSAF}. The game is challenging in that the verse and chorus section may repeat multiple times in a song, with sometimes minor variations. 
The boundaries are clear, and we use $n=10$ fragments (sections) per song.
In audio engineering, the task of arranging sections in a sensible way is referred to as \textbf{music sequencing}.




Lastly, we consider the \textbf{music medley} game, which aims to  put together a bag of short clips from different songs to build a longer music piece \cite{lin15tomm}. 
As the fragments (clips) are from different songs, the boundaries are also clear.
This is different from the cross-song puzzle considered in \cite{smith2017crosssong}. In music medley, we take \emph{one} fragment per song from $m$ ($=n$) songs, and aim to create \emph{an} ordering of them. In contrast, in cross-song puzzle, we take $\lfloor n/m \rfloor$ fragments per song from $m$ ($\neq n$) songs and aim to discern the origin of the fragments and get $m$ orderings.


\subsection{Problem Formulation}
All the aforementioned games are about ordering things. While solving an image jigsaw puzzle game, human beings usually consider the structural patterns and texture information as cues by comparing the puzzle pieces one by one \cite{noroozi2016unsupervised}. There is no need to put all the pieces in correct order all at once. As the number of permutations grows exponentially with $n$, we formulate the learning problem as a binary classification problem and predict whether a given pair of fragments is consecutive and is in correct order. 

In the training stage, all the fragments are segmented consecutively without overlaps per song, as shown in the leftmost part of Figure \ref{fig:flow}.
For each song, we get a collection of fragments $\{R_{1},\cdots,R_{n}\}$, which are in the correct order. Among the $2{{n}\choose{2}}$ possible fragments pairs, $n-1$ of them are in the correct order and are considered as the \textbf{positive data}, $\mathcal{P}_+ = \{(R_{i}, R_{i+1}) \mid i\in\{1,2,\dots,n-1\}\}$. 
While all the other possible pairs can be considered as the \textbf{negative data}, we consider only three types of them:
\[\left\{
\begin{array}{ll}
\mathcal{P}_-^{(1)} = \{(R_{i+1}, R_{i}) \mid i\in\{1,\dots,n-1\}\} \,,\\ 
\mathcal{P}_-^{(2)} = \{(R_{i}, R_{i+2}) \mid i\in\{1,\dots,n-2\}\} \,,\\
\mathcal{P}_-^{(3)} = \{(R_{i+2}, R_{i}) \mid i\in\{1,\dots,n-2\}\} \,.\\
\end{array}
\right. \]
Pairs of the first type is consecutive but in incorrect order. Pairs of the  second and third types are not consecutive. The negative data is the union of them: $\mathcal{P}_- = \mathcal{P}_-^{(1)} \cup \mathcal{P}_-^{(2)} \cup \mathcal{P}_-^{(3)}$. 
Therefore, the ratio of positive and negative data $|\mathcal{P}_+|/|\mathcal{P}_-|$ is about $1/3$.
In our experiments, we also refer to data pairs belonging to $\mathcal{P}_+$, $\mathcal{P}_-^{(1)}$, $\mathcal{P}_-^{(2)}$ and $\mathcal{P}_-^{(3)}$ as `R1R2,' `R2R1,' `R1R3' and `R3R1,' respectively.\footnote{We note that, in related work working on videos \cite{misra2016shuffle,fernando2016self,lee2017unsupervised}, they treated R1R2 the same as R2R1, and likewise R1R2R3 the same as R3R2R1, assuming that playing a short video clip in the reverse order is fine.}


\begin{figure*}[tb]
\centering
 \includegraphics[width=\textwidth]{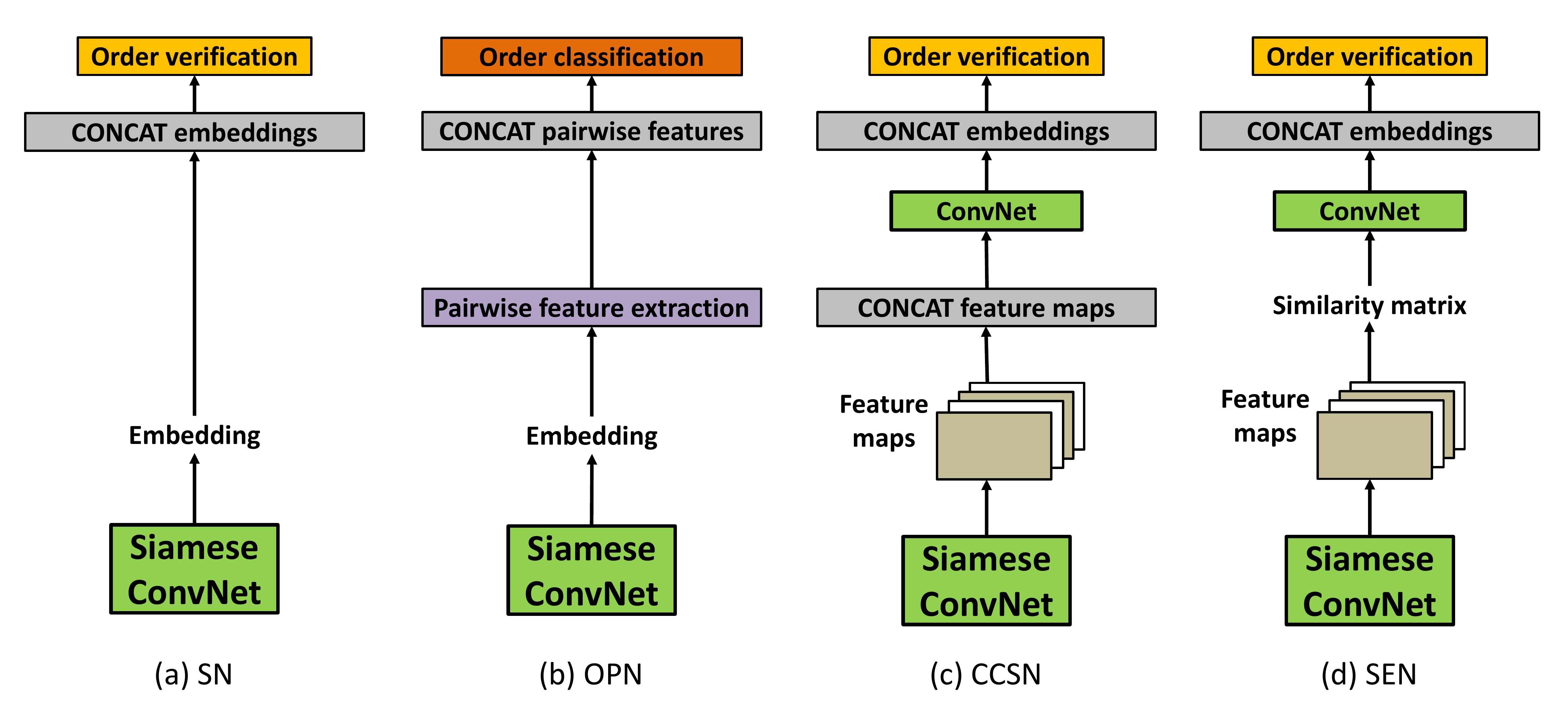}
\caption{Four different network architectures used in our experiments: (a) A standard Siamese network \cite{misra2016shuffle} for order verification (i.e. binary classification); (b) the order prediction network \cite{lee2017unsupervised}, which extracts pairwise features from each embedding and concatenates all pairwise features for order classification (i.e. multiclass classification); (c) the concatenated-convolutions Siamese network, which captures joint frames by concatenating all feature maps; and (d) the proposed similarity embedding network, which learns structural patterns from the similarity matrices.}
\label{fig:arch}
\end{figure*}

Given a training set $\mathcal{D} = \{(X,~y)~|~X\in \mathcal{P}, y \in \{0,1\}\}$, where $\mathcal{P}$ is the union of $\mathcal{P}_+$ and $\mathcal{P}_-$ from all the songs and $y$ whether a pair is positive or not, we learn the parameters $\theta$ of a neural network $f_\theta$ by solving:
\begin{equation}
\min_{\theta} \sum_{(X,y) \in \mathcal{D}}  \mathcal{L}(f_\theta(X),y) + \mathcal{R(\theta)} \\,
\end{equation}
where $\mathcal{L}$ is a loss function (e.g. cross entropy) and $\mathcal{R(\theta)}$ is a regularization term for avoiding overfitting.


\subsubsection{Global Ordering}
Given a data pair $X=(R_a,R_b), a,b\in\{1,\dots,n\}, a\neq b$, the estimate $f_\theta(X)$ is a value in $[0,1]$, due to a softmax function. For each song in the validation set, we need to get this estimate for all the data pairs, and seek to find the correct global ordering of the fragments from these estimates.\footnote{Our model is trained by playing only the music jigsaw puzzle, but in testing time the model will be applied to different games.}
While there may be other sophisticated ways doing it, we find the following simple heuristic works quite well already: we evaluate the ``fitness'' of any ordering of the fragments by summing the model output of the composing $n-1$ consecutive pairs. For example, the fitness for $(R_a,R_b,R_c)$, for $n=3$, will be $f_\theta(R_a,R_b)+f_\theta(R_b,R_c)$. We then simply pick the ordering with the highest fitness score as our solution for the game for that song.




\section{Network Architecture}

\subsection{Similarity Embedding Network (SEN)}

A Siamese network \cite{bromley94signatureverification} is composed of two (or more) twin subnetworks that share the same parameters. 
The subnetworks usually use convolutional layers (but there are exceptions \cite{mueller16aaai}). The outputs of the last convolutional layer are concatenated and then feed to the subsequent fully-connected layers. The functions of the convolutional layers and the fully connected layers are feature learning and classifier training, respectively. 
Because Siamese networks can process multiple inputs at the same time, it is widely used in various metric learning  problems \cite{chopra2005learning}.


As shown in the middle of Figure \ref{fig:flow}, the proposed SEN model also uses a convolutional Siamese network (Siamese ConvNet) to learn features from spectrogram-like 2D features of a pair of fragments.
However, motivated by a recent work \cite{luo2016efficient}, which used a product layer to compute the inner product between two representations of a Siamese network, we propose to compute the similarity matrix from the frame-by-frame output of the last layer of the Siamese ConvNet, and further learn features from the similarity matrix with a few more convoltutional layers, as shown in the right hand side of Figure \ref{fig:flow} (and Figure \ref{fig:arch}(d)). The output can be viewed as an embedding of the similarity matrix, therefore the name of the network. 

Given the output feature maps of the Siamese ConvNet, $\mathbf{G}_a = h_\theta(R_a) \in \mathcal{R}^{N \times k}$, $\mathbf{G}_b = h_\theta(R_b) \in \mathcal{R}^{M \times k}$, where $h_\theta$ denotes the network up to the last layer of the Siamese ConvNet, $N$ and $M$ the (temporal) length of the output and $k$ the dimension of the feature, the similarity matrix $\mathbf{S}\in \mathcal{R}^{N \times M}$ is computed by the Cosine score:
\begin{equation}
\mathbf{S}_{ij} = \left(\mathbf{g}_{a,i}^T \mathbf{g}_{b,j} \right) / \left(\|\mathbf{g}_{a,i} \|^2_2\| \mathbf{g}_{b,j} \|^2_2 \right) \,,
\label{eq:sim}
\end{equation}
where $\mathbf{g}_{a,i} \in \mathcal{R}^{k}$ is the $i$-th  feature (slice) of $\mathbf{G}_a$. 

Because we want the resulting similarity matrix to capture the temporal correspondence between the input fragments, in SEN we use 1D convolutions along the temporal dimension \cite{liu2016event} for the Siamese ConvNet. Moreover, we set the stride size to 1 and use no pooling layers in the Siamese ConvNet for SEN, to capture detailed temporal information of the fragments.


\subsection{Baselines}


\subsubsection{Siamese CNN (SN)}
A valid baseline is the pairwise Siamese ConvNet, which takes the input fragment pairs and learns a binary classifier for order verification.

\subsubsection{Concatenated-inputs CNN (CIN)}
An intuitive solver for the music jigsaw puzzle is to concatenate the spectrogram-like 2D features of the fragments along the time dimension, and use a CNN (instead of an SN) for order verification.
We suppose this model can catch the weird boundary of an incorrectly ordered fragment pair.

\subsubsection{Concatenated-convolutions Siamese Network (CCSN)}
This is a state-of-the-art network for image feature learning \cite{wang2016joint}. Given the feature maps from  the last convolutional layers of a Siamese ConvNet, we can simply concatenate them along the depth dimension (instead of computing the similarity matrix) and then use another stack of convolutional layers to learn features. As shown in Figure \ref{fig:arch}(c), the only difference between CCSN and SEN lies in how we extract information from the Siamese ConvNet.

\subsubsection{Triplet Siamese Network (TSN) \& Order Prediction Network (OPN)}
The state-of-the-art algorithms in solving video puzzle games use a \emph{list-wise} approach instead of a \emph{pair-wise} approach. The TSN model \cite{misra2016shuffle} is simply an expansion of SN by taking three inputs instead of two. 
In contrast, the OPN model \cite{lee2017unsupervised}, depicted in Figure \ref{fig:arch}(b), takes all the $n$ fragments at the same time,
aggregates the features from all possible feature pairs for feature learning, and seeks to pick the best global ordering out of the $n!$ possible combinations via a multiclass classification problem. 

\subsection{Implementation Details}

As done in many previous works \cite{dieleman2014end}, we compute the spectrograms by sampling the songs at 22,050 Hz and using a Hamming window of 2,048 samples and hop size 512 samples. We then transform the spectrograms into 128-bin log mel-scaled spectrograms and use that as input to the networks, after z-score normalization. 

Unless otherwise specified, in our implementation all the Siamese ConvNets use 1D convolutional filters (along the time dimension), with the number of filters being 128, 256, 512, respectively, and the filter length being 4. For SEN and CCSN, the convolutional filters for the subsequent ConvNet are 64, 128, 256, respectively, followed by 3 by 3 maximum pooling and the filter size is also 3 by 3. Here, SEN uses 2D convolutions, while CCSN uses 1D convolutions. 
Except for TSN and OPN, we use a global pooling layer (which is written as `CONCAT' in Figure \ref{fig:arch}) after the ConvNet in SEN, CCSN, CIN, and the Siamese ConvNet in SN. 
The dimension of the two fully-connected layers after this pooling layer are all set to 1,024. 
All networks use rectified linear unit (ReLU) as the activation function everywhere. 
Lastly, all the models are trained using stochastic gradient descent with momentum 0.9, with batch size setting to 16.


\begin{table}
\centering
\caption{The pairwise accuracy and global accuracy on $n=3$ fixed-length jigsaw puzzle.}
\label{table:all}
\begin{tabular}{|l|cc|}
\hline
\multirow{ 2}{*}{Method} &pairwise   &global \\
&accuracy&accuracy\\
\hline\hline
SN  &0.851  &0.825 \\
CCSN \cite{wang2016joint} &0.872  &0.840\\
CIN  &0.912  &0.864\\
TSN \cite{misra2016shuffle} &0.911  &0.890\\
OPN \cite{lee2017unsupervised} &0.929  &0.916\\
\hline
SEN (proposed) &{0.996}  &{0.994}\\
\hline
\end{tabular}
\end{table}

\section{Experiments}
\subsection{Data sets}
Any music collection can be used in our puzzle games, since we do not need any human annotations. In this paper, 
we use an in-house collection of 31,377 clips of Pop music as our corpus. 
All these clips are audio previews downloaded from the Internet, with unknown starting point in each song the audio preview was extracted from. 
All these clips are longer than 24 seconds, so we consider only  the first 24 seconds per clip for simplicity of the model training process.
Moreover, we randomly pick 6,000 songs as validation set, 6,000 songs for testing, and the remaining 19,377 clips for training.

Different data sets are used as the test set for different music puzzle games. 
For \textbf{music jigsaw puzzle}, we simply use the test set of the in-house collection.
For \textbf{music sequencing}, we use the popular music subset of the RWC database \cite{goto2002rwc}, which contains 100 complete songs with manually labeled section boundaries \cite{goto2006aist}.\footnote{\url{https://staff.aist.go.jp/m.goto/RWC-MDB/AIST-Annotation/}} We divide songs into fragments according to these boundaries. The number of resulting fragments per song ranges from 11 to 34. For simplicity, we consider only the first ten fragments (sections) per song.
For \textbf{music medley}, we collect 16 professionally-compiled medleys of pop music (by human experts) from YouTube. Each medley contains 7 to 11 different short music clips, whose length vary from 5 to 30 seconds. For reproducibility, we have posted the YouTube links of these medleys in our project website.

We need to perform downbeat tracking for the in-house collection. To this end, we use the implementation of a state-of-the-art recurrent neural network available in the Python library madmom\footnote{\url{https://github.com/CPJKU/madmom}} \cite{bock2016madmom}. After getting the downbeat positions, we randomly choose some of them so that each fragment is about $\lfloor24/n \rfloor$ seconds in length.


\subsection{Result on 3-Piece Fixed-length Jigsaw Puzzle}
As the first experiment, we consider the $n=3$ jigsaw puzzle game, using 1,000 clips randomly selected from the test set of the in-house collection. 
Accordingly, we train all the neural networks (including the baselines) by playing $n=3$ jigsaw puzzles using the training set of the same data set. 
All the clips are segmented at equally-spaced timepoints. Therefore, the length of fragments is fixed to 8 seconds. This corresponds to the simplest setting in Table \ref{table:aspect}.

We employ the \textbf{pairwise accuracy} (PA) and \textbf{global accuracy} (GA) as our performance metrics. For an ordering of $n$ fragments, GA requires it to be exactly the same as the groundtruth one, whereas PA takes the average of the correctness of the $n-1$ composing pairs. 
For example, ordering ($R_{1}$,$R_{2}$,$R_{3}$) as ($R_{2}$,$R_{3}$,$R_{1}$) would get 0.5 PA (for the pair ($R_{2}$,$R_{3}$) is correct) and 0 GA.

The result is shown in Table \ref{table:all}. The performance of the baseline models seem to correlate well with their sophistication, with SN performing the worst (0.825 GA) and OPN \cite{lee2017unsupervised} performing the best (0.916 GA). The comparison between SN and TSN \cite{misra2016shuffle} implies that more inputs offers more information. Moreover, the results in GA and PA seem to be highly correlated.

More importantly, by comparing the result of SEN against the baseline models, we see that SEN outperforms the others by a great margin, reaching almost 100\% accuracy for both metrics.  In particular, the performance gap between CCSN and SEN suggests that learning from the similarity matrix seems to be an important design.




\begin{table}[t]
\centering
\caption{The accuracy on music jigsaw puzzles with different segmentation method (fixed-length or downbeat informed) and different number of fragments. Pairwise accuracy is outside the brackets and global accuracy is inside.}
\label{table:3468}
\begin{tabular}{|cc|ccc|}
\hline
& $n$  &SN  &CIN  &SEN\\
\hline\hline
\parbox[t]{2mm}{\multirow{4}{*}{\rotatebox[origin=c]{90}{fixed}}}
&3  &0.851 (0.825)  &0.912 (0.864)  &{0.996 (0.994)}\\
&4  &0.752 (0.641)  &0.844 (0.722)  &{0.990 (0.982)}\\
&6  &0.609 (0.304)  &0.761 (0.455)  &{0.989 (0.977)}\\
&8  &0.514 (0.110)  &0.682 (0.229)  &{0.985 (0.953)}\\
\hline\hline
\parbox[t]{2mm}{\multirow{4}{*}{\rotatebox[origin=c]{90}{downbeat}}}
&3  &0.781 (0.692)  &0.912 (0.863)  &{0.995 (0.991)}\\
&4  &0.681 (0.472)  &0.871 (0.761)  &{0.995 (0.987)}\\
&6  &0.536 (0.171)  &0.764 (0.499)  &{0.991 (0.971)}\\
&8  &0.449 (0.056)  &0.680 (0.297)  &{0.990 (0.961)}\\
\hline
\end{tabular}
\end{table}

\subsection{Result on Variable-length Jigsaw Puzzle}

Next, we consider music jigsaw puzzles with variable-length fragments. Thanks to the global pooling layer, in our implementation SEN, SN and CIN can take input of arbitrary length, even if the length of the training fragments is different from the length of the testing fragments. Moreover, the pair-wise strategy allows these three models to tackle $n>4$ puzzle games, while some other models such as OPN cannot. Therefore, we only consider SEN, SN and CIN in the following experiments.


We create different games by varying $n$ from 3, 4, 6 to 8, using the uniformly segmented fragments from the in-house data set. The length of the fragments is hence 8, 6, 4, 3 seconds, respectively. 
Among them, the $n=8$ game is the most challenging one, partly due to the number of fragments and partly due to their shorter duration (implying less information per fragment).
We use the same SEN, SN and CIN models trained by solving $n=3$ jigsaw puzzles. 
In addition, we aim at comparing the result of using different segmentation methods to process both the training and test clips. Therefore, we re-train the SEN, SN and CIN models trained by solving $n=3$ jigsaw puzzles segmented at downbeat positions, and apply them to also downbeat-informed jigsaw puzzles with different values of $n$.


\begin{table}[t]
\centering
\caption{The accuracy of SEN on three kinds of puzzle game for two segmentation methods.}
\label{table:fixdown}
\begin{tabular}{|l|cc|}
\hline
game  &fixed-length  &downbeat-informed\\
\hline
\hline
puzzle ($n=8$)  &0.985 (0.953)  &{0.990 (0.961)}\\
sequencing  &0.789 (0.440)  &{0.937 (0.790)}\\
medley  &0.945 (0.688)  &{0.961 (0.750)}\\
\hline
\end{tabular}
\end{table}

\begin{table*}[tb]
\centering
\caption{The result of a few ablated version of SEN for different music puzzle games.}
\label{table:ablationnegative}
\begin{tabular}{|l|cccc|ccc|c|}
\hline
game  &\begin{tabular}{@{}c@{}}Inner \\ product\end{tabular}  &\begin{tabular}{@{}c@{}}Conv \\ stride 2\end{tabular}  &\begin{tabular}{@{}c@{}}Global P \\ (mean)\end{tabular}  &\begin{tabular}{@{}c@{}}Global P \\ (max)\end{tabular}  &\begin{tabular}{@{}c@{}}R2R1 \\ only\end{tabular}  &\begin{tabular}{@{}c@{}}R1R3 \\ only\end{tabular}
 &\begin{tabular}{@{}c@{}}R3R1 \\ only\end{tabular}  &All\\
\hline
\hline
puzzle  &0.90 (0.69)  &0.65 (0.17)  &0.96 (0.87)  &0.98 (0.93)  &0.84 (0.57)  &0.97 (0.87)  &0.96 (0.86)  &0.99 (0.96)\\
sequencing  &0.74 (0.38)  &0.54 (0.06)  &0.81 (0.49)  &0.92 (0.76)  &0.62 (0.22)  &0.81 (0.46)  &0.91 (0.69)  &0.94 (0.79)\\
medley  &0.88 (0.50)  &0.73 (0.13)  &0.81 (0.56)  &0.93 (0.69)  &0.86 (0.44)  &0.93 (0.69)  &0.90 (0.63)  &0.96 (0.75)\\
\hline
\end{tabular}
\end{table*}

Table \ref{table:3468} shows the result. We can see that the result of SN and CIN both decrease quite remarkably as the value of $n$ increases, and that the downbeat-informed games are indeed slightly more challenging than the fixed-length games, possibly due to the clarity at the boundary. When $n=8$, the PA and GA of SN drop to 0.514 and 0.110, whereas the PA and GA of CIN drop to 0.682 and 0.229. However, the accuracy of the SEN model remains high even when $n=8$ (0.985 PA and 0.953 GA), suggesting that SEN can work quite robustly against various music jigsaw puzzles.



\subsection{Result on Music Sequencing and Music Medley}

Lastly, we evaluate the performance of SEN on music sequencing and music medley, which are supposed to be more challenging than jigsaw puzzles. We do not consider SN and CIN here, for their demonstrated poor performance in $n=8$ jigsaw puzzles. Instead, we compare two SEN models, one trained with uniform segmentation ($n=3$) and the other with downbeat-informed segmentation ($n=3$). 


From Table \ref{table:fixdown}, we can see that these two games are indeed more challenging than jigsaw puzzles. When using a SEN model trained with uniform segmentation, the GA can drop to as low as 0.440 for music sequencing and 0.688 for music medley. However, more robust result can be obtained by training SEN using downbeat-informed segmentation: the GA would be improved to 0.790 and 0.750 for the two games, respectively. 
This is possibly because the downbeat-informed segmentation can avoid SEN from learning only low-level features at the  boundary of fragments.

\begin{figure*}[tb]
\centering
\begin{tabular}{@{}ccc@{}}
        \includegraphics[width=.32\textwidth]{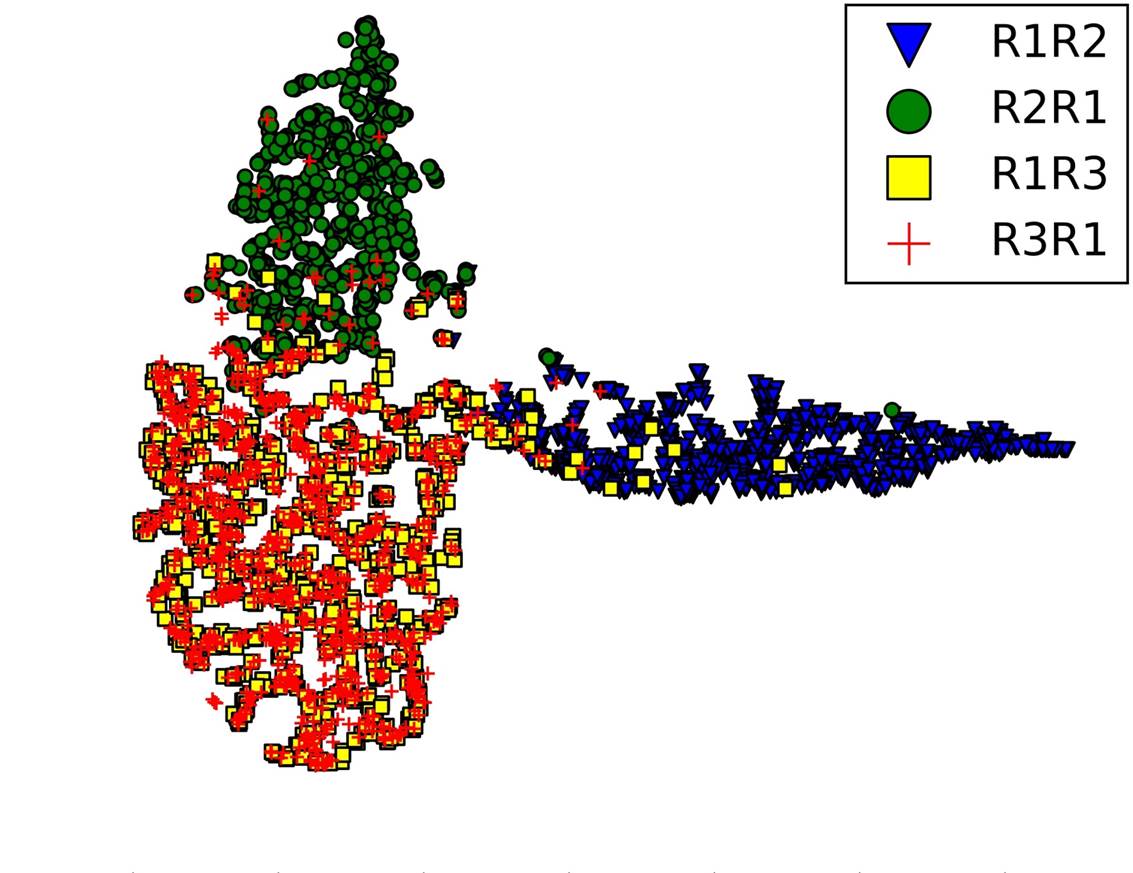}
        \includegraphics[width=.32\textwidth]{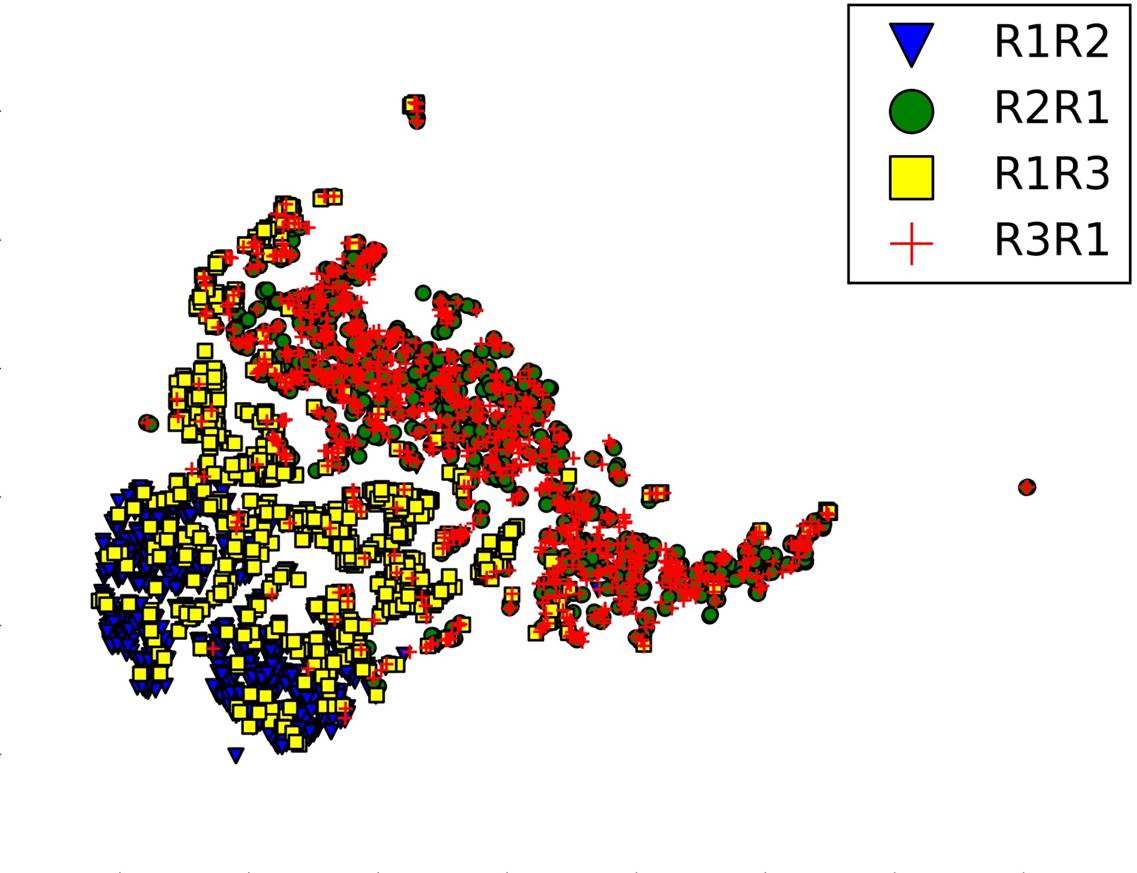}
        \includegraphics[width=.32\textwidth]{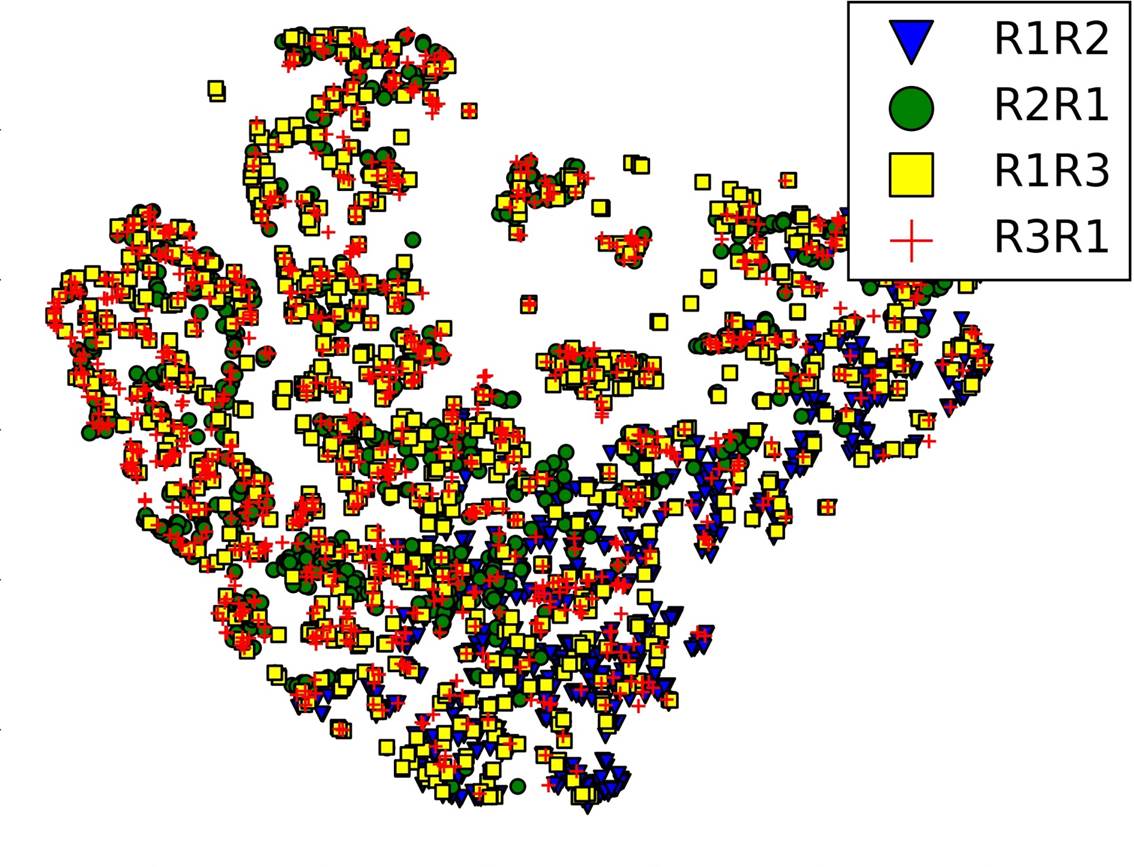}
\end{tabular}
\caption{Embeddings of different data pairs learned by (from left to right) SEN, CCSN and SN, respectively. The embeddings are projected to a 2D space for visualization via t-SNE \cite{maaten2008visualizing}. The figure is best viewed in color.}
\label{fig:embedding}
\end{figure*}

\begin{figure}[tb]
 \includegraphics[width=.45\textwidth]{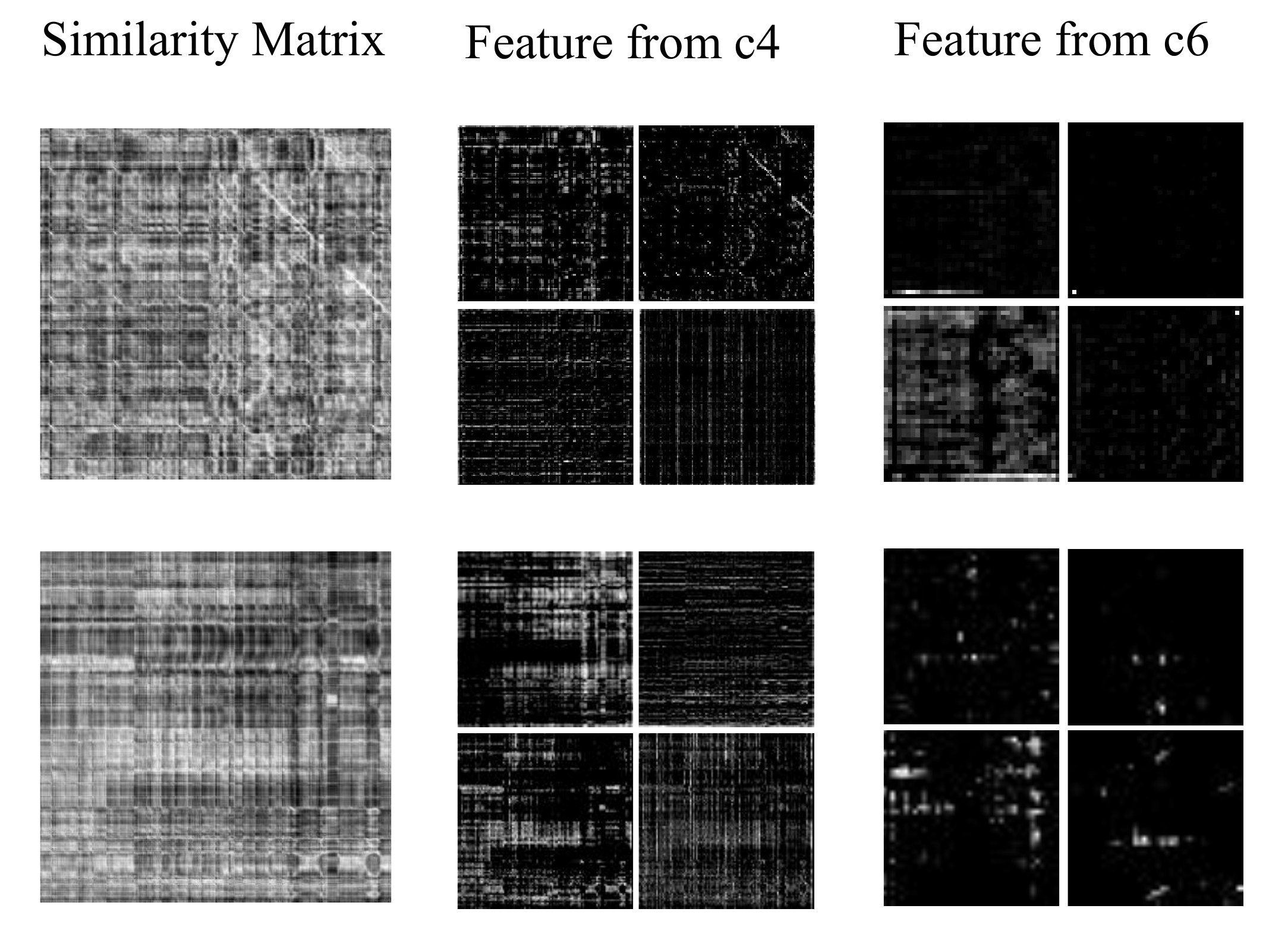}
\centering
\caption{Visualizations of the features learned from the c4 and c6 layers of SEN for two pairs of fragments.}
\label{fig:visual}
\end{figure}


We perform an error analysis looking into the incorrect prediction in the music sequencing game, which has some musically meaningful insights. A song is composed of several sections, such as intro (\texttt{I}), verse (\texttt{V}), chorus (\texttt{C}) and bridge (\texttt{B}), with some variations such as \texttt{Va} and \texttt{Vb}. A correct global ordering of one of the songs in RWC is: \texttt{I}-\texttt{Va}-\texttt{Ba}-\texttt{Vb}-\texttt{Cpre}-\texttt{Ca}-\texttt{Bb}-\texttt{Va}-\texttt{Vc}-\texttt{Cpre}. For this song, the estimated ordering of SEN is: \texttt{I}-\texttt{Bb}-\texttt{Va}-\texttt{Vc}-\texttt{Cpre}-\texttt{Ca}-\texttt{Va}-\texttt{Ba}-\texttt{Vb}-\texttt{Cpre}. We can use a numerical notation and represent our result as \texttt{1}-\texttt{7}-\texttt{8}-\texttt{9}-\texttt{10}-\texttt{6}-\texttt{2}-\texttt{3}-\texttt{4}-\texttt{5}. 
We can see that the local prediction of \texttt{7}-\texttt{8}-\texttt{9}-\texttt{10} and \texttt{2}-\texttt{3}-\texttt{4}-\texttt{5} is in correct order. 
Moreover, these two passages are fairly similar (both have the structure \texttt{V}-\texttt{Cpre}). 
Therefore, the predicted ordering may sound right as well. 
Indeed, we found that most of the incorrect predictions are correct in local ordering.

We use user-created medleys as the groundtruth in the medley game, but as the creation of a medley is an art, different orderings may sound right. Therefore, we encourage readers to visit our project website to listen to the result.

\subsection{Ablation Analysis}
We assess the effect of various design of downbeat-informed SEN by evaluating ablated versions. Table \ref{table:ablationnegative} shows the result when we (from left to right): i) replace cosine similarity in Eq. (\ref{eq:sim}) by inner product, ii) increase the stride of the convolutions in Siamese ConvNet from 1 to 2, iii) use only global mean pooling or global max pooling (we use the concatenation of mean, max and standard deviation in our full model), and iv) use one type of negative data only.
Most of these changes decrease the accuracy of SEN. Some observations:
\begin{itemize}
\item Calculating the similarity matrix using the inner product cannot guarantee that the similarity scores are in the range of $[0,1]$ and this hurts the accuracy of SEN. 
\item Setting the stride size larger can speed up the training process, but doing so losses much temporal information. 
\item Max pooling alone works quite well for the global pooling layer, but it is even better to also consider mean and standard deviation.
\item Using R2R1 as the negative data alone is far from sufficient. Actually, both R1R3 only and R3R1 only seem to work better than R2R1 only. The best result (especially in GA) is obtained by using all three types of negative pairs. 
\end{itemize}

\subsection{What the SEN Model Learns?}
Figure \ref{fig:embedding} shows the embeddings (output of the last fully-connected layer) of different data pairs learned by SEN, CCSN and SN, respectively.
We can clearly see that the positive and negative pairs can be fairly easily distinguished by the embeddings learned by SEN. Moreover, SEN can even distinguish R2R1 (consecutive) from R1R3 and R3R1 (non-consecutive). 
This is an evidence of the effectiveness of SEN in learning sequential structural patterns.



Finally, Figure \ref{fig:visual} shows the features from the first (c4) and last convolution (c6) layers in the ConvNet of SEN, given two randomly chosen pairs (the first row is R1R2 and the second row is R1R3).  We see that the filters detect different patterns and textures from the similarity matrices. 

\section{Conclusion}
In this paper, we have presented a novel Siamese network called the similarity embedding network (SEN) for learning sequential patterns in a self-supervised way from similarity matrices. We have also demonstrated the superiority of SEN over existing Siamese networks using different types of music puzzle games, including music medley generation.

In our evaluation, however, music medley generation is viewed as just one of the evaluation tasks. In future work, we will focus more on the music medley generation task itself. For example, a listening test should be conducted for subjective evaluation.
We also plan to deeply investigate the features or patterns our model learns for generating medleys and correlate our findings with those reported in related work on automatic music mashup \cite{davies2014automashupper} and playlist sequencing \cite{bittnerautomatic}. We also want to investigate thumbnailing methods \cite{huang17apsipa,kim17ismirlbd} to pick fragments from different songs, and methods such as beat-match and cross-fade \cite{bittnerautomatic} to improve the transition between clips.

\bibliographystyle{aaai} 
\bibliography{refs}
\end{document}